\documentclass{article}

\usepackage{microtype}
\usepackage{graphicx}
\usepackage{subfigure}
\usepackage{booktabs} 

\usepackage{hyperref}

\usepackage{multirow}
\usepackage[table]{xcolor}
\usepackage{tabularx}
\usepackage{adjustbox}

\usepackage[accepted]{icml2025}

\usepackage{amsmath}
\usepackage{amssymb}
\usepackage{mathtools}
\usepackage{amsthm}

\usepackage[capitalize,noabbrev]{cleveref}

\theoremstyle{plain}

\theoremstyle{definition}

\theoremstyle{remark}

\usepackage[textsize=tiny]{todonotes}
\definecolor{mycolor}{rgb}{0.925,0.957,1.000}
\definecolor{mycolor1}{rgb}{1,0.8,0.8}
\definecolor{mycolor2}{rgb}{1,0.8,0.9}

\icmltitlerunning{MINT: Mitigating Hallucinations in Large Vision-Language Models via Token Reduction}

\begin{document}

\twocolumn[
\icmltitle{MINT: Mitigating Hallucinations in Large Vision-Language Models via Token Reduction}



\icmlsetsymbol{equal}{*}
\icmlsetsymbol{corresponding}{$\dagger$}

\begin{icmlauthorlist}
\icmlauthor{Chao Wang}{ai,ins,corresponding}
\icmlauthor{Jianming Yang}{ai,ins}
\icmlauthor{Yang Zhou}{cs,corresponding}
\end{icmlauthorlist}
\icmlaffiliation{ai}{School of Future Technology, Shanghai University, Shanghai, China.}
\icmlaffiliation{ins}{Institute of Artificial Intelligence, Shanghai University, Shanghai, China.}
\icmlaffiliation{cs}{School of Computer Engineering and Science, Shanghai, China}

\icmlcorrespondingauthor{Chao Wang}{cwang@shu.edu.cn}
\icmlcorrespondingauthor{Yang Zhou}{saber\_mio@shu.edu.cn}

\icmlkeywords{Machine Learning, ICML}

\vskip 0.3in
]



\printAffiliationsAndNotice{\icmlEqualContribution} 

\begin{abstract}
Hallucination has been a long-standing and inevitable problem that hinders the application of Large Vision-Language Models (LVLMs) in domains that require high reliability. Various methods focus on improvement depending on data annotations or training strategies, yet place less emphasis on LLM's inherent problems. To fill this gap, we delve into the attention mechanism of the decoding process in the LVLM. Intriguingly, our investigation uncovers the prevalent attention redundancy within the hierarchical architecture of the LVLM, manifesting as overextended image processing in deep layers and an overabundance of non-essential image tokens. Stemming from the observation, we thus propose MINT, a novel training-free decoding strategy, \textbf{MI}tigating hallucinations via toke\textbf{N} reduc\textbf{T}ion. Specifically, we dynamically intensify the LVLM's local perception capability by masking its attention to irrelevant image tokens. In addition, we use contrastive decoding that pushes the model to focus more on those key image regions. Our full method aims to guide the model in concentrating more on key visual elements during generation. Extensive experimental results on several popular public benchmarks show that our approach achieves a 4\% improvement in mitigating hallucinations caused by distracted perception compared to original models. Meanwhile, our approach is demonstrated to make the model perceive 5\% more visual points even though we reduce a suite of image tokens. 
\end{abstract}

\section{Introduction}\label{introduction}
Benefiting from those advanced large language models (LLMs) \cite{touvron2023llama, vicuna2023, bai2023qwen}, large Vision-Language models (LVLMs) have made significant strides \cite{liu2024improved-llava, bai2023qwen-vl}, exhibiting a superior ability to generate contextually coherent responses. Generally, those models bridge the semantic gap between linguistic instructions and visual content by aligning the image features from vision encoders and LLMs, harnessing LLMs' superior text generation capability. Despite ongoing advancements, the persistent challenge of hallucinations continues to hinder the practical applications (e.g., autonomous driving \cite{wen2023dilu}, medical VQA \cite{yan2024medicalvqa}, etc.) of LVLMs, where high reliability is crucial in meeting demands.

\begin{figure*}[t]
\vskip 0.1in
\centering
\includegraphics[width=0.95\textwidth]{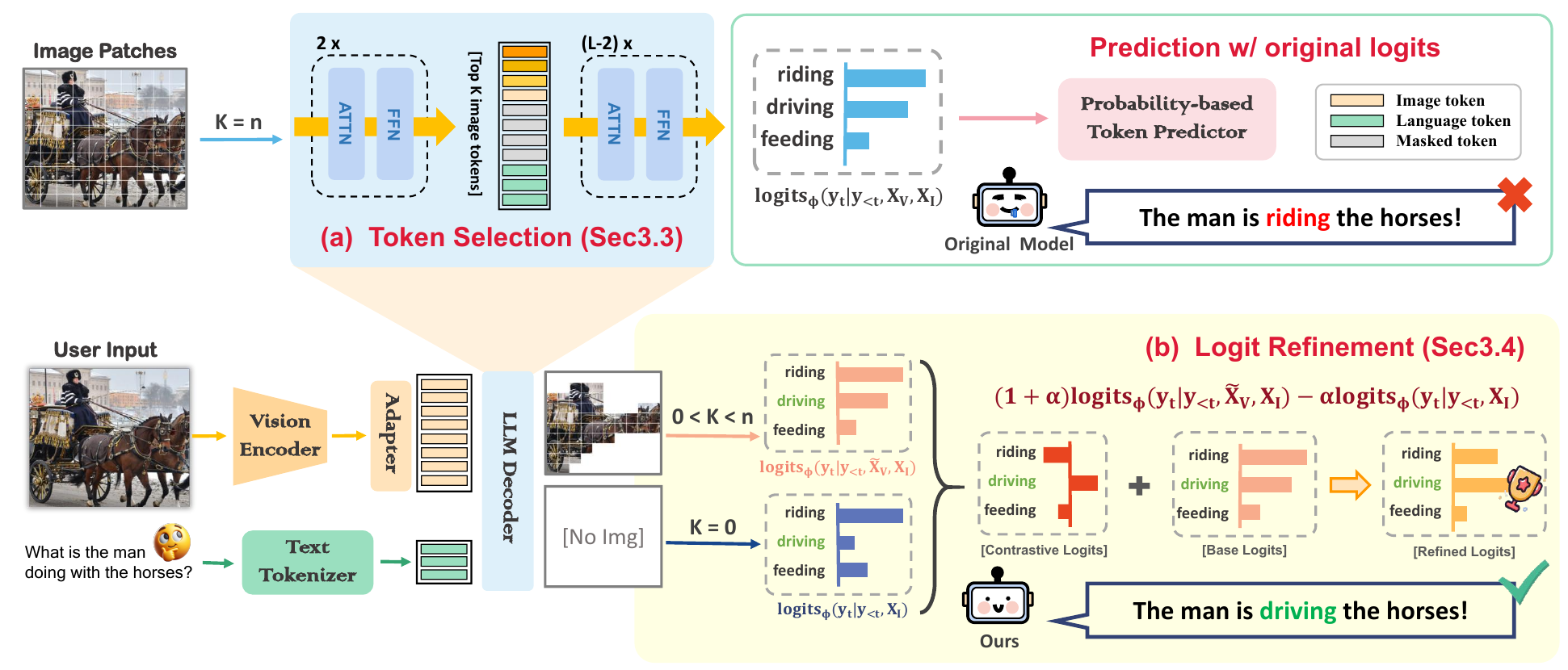}
\caption{\textbf{The overview of our method MINT.} We first add a token selection module inside the LLM Decoder to enable control of the model's dynamic concentrating on key image regions. Subsequently, we use contrastive decoding to amplify the impact of the selected image tokens, offering reference for logit refinement. The final calibrated logit can lead to more reliable token prediction for improved model performance.}
\label{fig:overview}
\end{figure*}

As the push for more reliable generative AI intensifies, LVLMs have been polished up in many aspects including data diversity \cite{gunjal2024fdpo, wu2024q-instruct}, training strategies \cite{hu2022lora, liu2024llava, yu2024rlhf}, model structures \cite{liu2024llavanext, dong2024internlmxcomposerhd}, etc., substantially mitigating their hallucinations and enhancing their performance. However, except for these aspects that need resource-intensive retraining, the stubborn bias \cite{leng2024vcd, liu2025pai} that LVLMs inherit from LLMs is often overlooked. Specifically, the existing generative paradigms mapping visual representations into the language representation space lead to the dominant status of LLMs. Therefore, these image tokens are always treated as language tokens by LLMs, resulting in the model ignoring the earlier image tokens and even all of them just like LLMs do \cite{xiao2023streamllm} during the generation. 

Drawing inspiration from previous studies \cite{chen2025fastv, liu2025pai}, we delve into the attention mechanism of the Transformer-based decoder within LVLMs to further recognize the next-token generation. We conduct an analysis of attention allocation during the LVLM's token generation, finding that \emph{the imbalanced attention allocation persists through all the decoding layers with significant differences between shallow and deep layers}. Specifically, the LVLM tends to finish comprehending the input image and user's question in the first two layers, while distributing redundant attention to useless tokens (e.g., system prompt) that do not carry any semantic information in the subsequent layers. Furthermore, we continue to visualize the attention heatmap in each layer, which highlights the attention intensities over image patches when the LVLM generates a specific token. Intriguingly, the observation clearly reveals that the LVLM has already built a rich understanding of the image in shallow layers, as displayed in Fig.\ref{fig:layer_pattern_comparison}. Similarly, we find that there is also attention redundancy when processing image tokens, implying that only a part of the key image tokens are necessary in each prediction.

Based on these observations, we thus propose our approach focusing on local perception enhancement, which dynamically selects key image tokens according to attention patterns in the shallow decoding layer. We then adopt contrastive decoding to push the LVLM to focus attention on these selected image tokens further. Extensive experiments demonstrate that our method indeed helps the model perceive more visual details while mitigating hallucinations. 
In brief, our main \textbf{contributions} are concluded as follows:

\textbf{1).} An empirical analysis of attention patterns in the decoder is conducted, revealing imbalanced attention allocation and significant redundancy in attention mechanisms within LVLMs, alongside diverse attention patterns specifically concentrated in the initial two decoding layers.

\textbf{2).} We observe that the model's early-stage perception of images can effectively guide its focus toward task-specific regions. Based on this insight, a novel training-free method, MINT, is proposed to mitigate hallucination in LVLMs without relying on any assistance from external models.

\textbf{3).} Comprehensive experiments across various benchmarks and evaluations, involving models of different sizes, validate the effectiveness and generalizability of our approach.

\section{Related Work}\label{related work}
\subsection{Large Vision-Language Models}
The existing boom of LVLMs \cite{liu2024improved-llava, bai2023qwen-vl, liu2024llavanext, dong2024internlmxcomposerhd} could be chiefly attributed to the tremendous success of LLMs technology, such as Llama \cite{touvron2023llama}, Vicuna \cite{vicuna2023}, Qwen \cite{bai2023qwen}, etc. These LVLMs try to build up a connection between the pre-trained LLMs and the off-the-shelf vision encoders \cite{dosovitskiy2020vit, radford2021clip}, utilizing a lightweight adapter, such as MLP \cite{liu2024improved-llava} or Qformer \cite{bai2023qwen-vl}, to translate visual features into language tokens understandable for LLMs. Therefore, they can perform next-token generation just like LLMs, iteratively predicting tokens based on produced probability distribution. 
Recent visual instruction tuning \cite{liu2024llava, liu2024improved-llava} has further enhanced LVLMs' ability to follow human's various instructions. Besides, a suite of works \cite{cha2024honeybee, yao2024deco} has also focused on improved projection modules to convert the aligned visual features effectively yet efficiently, retaining more detailed information contained in images for perception enhancement. Another series of studies \cite{lai2024lisa, zhang2025llava-grounded} has additionally explored equipping LVLMs with richer capabilities to conduct vision-centric tasks including referring object detection, reasoning segmentation, etc. Despite the ongoing evolution of LVLMs, it is still found that they are subject to hallucinations, generating content that is inconsistent with facts in images. 

\subsection{Hallucination Mitigation of LVLMs}
The phenomenon of hallucination in LVLMs \cite{fu2025blink} refers to the contradiction between the textual response and the factual information of the input image. Different origins are focused on, which we divide into \textit{imperfect alignment}, \textit{defective perception mechanism}, and \textit{over-reliance on language prior}. 

\textbf{Imperfect Alignment.} 
The most direct reason for hallucination in LVLMs is that there are still knowledge gaps between vision and language modalities. Therefore, high-quality curated datasets \cite{liu2024improved-llava, gunjal2024fdpo, wu2024q-instruct} and evolutionary fine-tuning methods \cite{hu2022lora, yu2024rlhf, wu2024reft} are introduced to further enrich the knowledge of LVLMs. 
For example, SpatialVLM \cite{chen2024spatialvlm} significantly enhances the model's spatial reasoning capability by training LVLMs with Internet-scale 3D spatial knowledge. Despite satisfactory improvement in a specific domain, these methods are mainly dependent on data-driven principles, ignoring the inherent problems of current LVLMs.

\textbf{Defective Perception Mechanism.}
As we know, the visual perception of LVLMs is achieved through integration with models like Vision Transformer (ViT) \cite{dosovitskiy2020vit} as the backbone visual encoders. However, biases such as high-norm outlier tokens are found in ViT \cite{darcet2024vit-flaw}, which could be abused in downstream tasks. Simultaneously, V* \cite{wu2024vstar} discovers severe visual information loss in high-resolution scenarios due to image compression of visual encoders. Thus, extra vision models are involved to detect task-relevant regions and feed them back to the LVLM. In contrast, LLaVA-Next \cite{liu2024llavanext} and InternLM-XComposer2-4KHD \cite{dong2024internlmxcomposerhd} achieve performance gains by treating each sub-region of the image as a single image, thereby increasing the image input resolution of the LVLMs. 

\textbf{Over-reliance on language prior.}
Another stubborn drawback of LVLMs is that they tend to generate textual responses directly depending on their constituent LLM's in-built knowledge, thus weakening the visual perception and exacerbating the occurrence of hallucination.
To highlight the impact of the image, VCD \cite{leng2024vcd} introduces contrastive decoding \cite{li2023cd} to utilize the discrepancy between two probability distributions, respectively, obtained from the original input image and a Gaussian blurred one, as the instructive value for probability refinement. In addition, ICD \cite{wang2024ICD} uses the prompt ``You are a confused object detector'' to get the distorted distribution. Differently, PAI \cite{liu2025pai} proposes increasing LVLMs' attention values of the corresponding image tokens to obtain a vision-centric distribution for contrastive decoding. Unlike PAI, our approach aims to dynamically make the LVLM focus on key regions.  

\section{Method}\label{method}
\subsection{Preliminaries}
In this work, we conduct our study based on several popular open-source LVLMs. The general structure of these LVLMs typically consists of an image encoder, an adapter for image features projection, and a pre-trained LLM as the language decoder. As the first step, the input image, divided into $n$ patches, would be transformed into $n$ tokens via the image encoder (e.g., Clip-ViT). The adapter subsequently projects these image tokens to the language representation space, matching the embedding size of the LLM and making visual information understandable for LLMs. We denote the $n$ image tokens as $X_{V} = \{{v}_{i}|0 \leq i < n\}$. The input textual prompt, tokenized into language tokens $X_{I}$, is then concatenated with the processed image tokens and concurrently fed into the LLM to generate responses.

The whole generation procedure encompasses a series of autoregressive next-token predictions. Each individual prediction can be formulated as follows:
\begin{equation}
{p}_{t} = \rho \left[ \mathrm{logits}_{\phi}({y}_{t}|{y}_{< t},{X}_{V},{X}_{I}) \right],
\end{equation}   
where $p_t$ is the overall probability distribution of all tokens in the model's vocabulary. $\rho\left[\cdot\right]$ represents the softmax function and $\mathrm{logits}_{\phi}(\cdot)$ represents the logit distribution obtained through the language decoder. $y_t$ denotes the current token to be predicted and ${y}_{< t}$ denotes the generated tokens from 0 to $t-1$ steps. $X_{I}=\{ X_{sys}, X_{que} \}$ refers to the input instruction including the system prompt (e.g., ``You are a helpful assistant.'') and the input question. The model then adopts a proper sampling strategy to select a token as the target one based on $p_t$. The generated token would be concatenated to the last of the history token sequence, serving as input in the next-round token prediction. The entire autoregressive token generation continues until an ``[EoS]'' (End of Sequence) token is produced or the number of the generated tokens has reached the maximum limit.

\subsection{Empirical Analysis} \label{empirical analysis}
To empirically investigate how LVLMs process information in visual tasks, we delve into the attention mechanism inside the Transformer-based decoder layers within the model. 

\textbf{Imbalanced Attention Allocation.}
Provided that the language decoder consists of $L$ layers, each with $H$ attention heads. For attention allocation of the $t$-th token in the $i$-th head of the $j$-th layer, we have:
\begin{equation}
a^{i,j}_{t,sys}+a^{i,j}_{t,img}+a^{i,j}_{t,que}+a^{i,j}_{out_{<t}} = 1,
\end{equation}
where $i\in[0,H-1]$ and $j\in[0,L-1]$. $a^{i,j}_{t,sys}$, $a^{i,j}_{t,img}$, $a^{i,j}_{t,que}$ and $a^{i,j}_{out_{<t}}$ respectively represents the total attention score the $t$-th token attends to tokens of different types including the system prompt, the input image, the input question, and generated output tokens. Therefore, we can compute the average attention score $\gamma$ that one type of tokens received in each attention head of the $j$-th layer as follows:
\begin{equation}\label{equa:average_attention}
    \gamma^j_{type}=\frac{1}{(N-1)\cdot H\cdot L} \sum_{t=1}^{N-1} {\sum_{i=0}^{H-1} {\sum_{j=0}^{L-1} a^{i,j}_{t,type}}},
\end{equation}
where $type\in\{sys, img, que, out\}$ and $N$ refers to the number of the generated tokens. Note that we exclude the first generated token since there is no output token before it.

\begin{figure}[htbp]
\vskip 0.1in
\centering
\includegraphics[width=\linewidth]{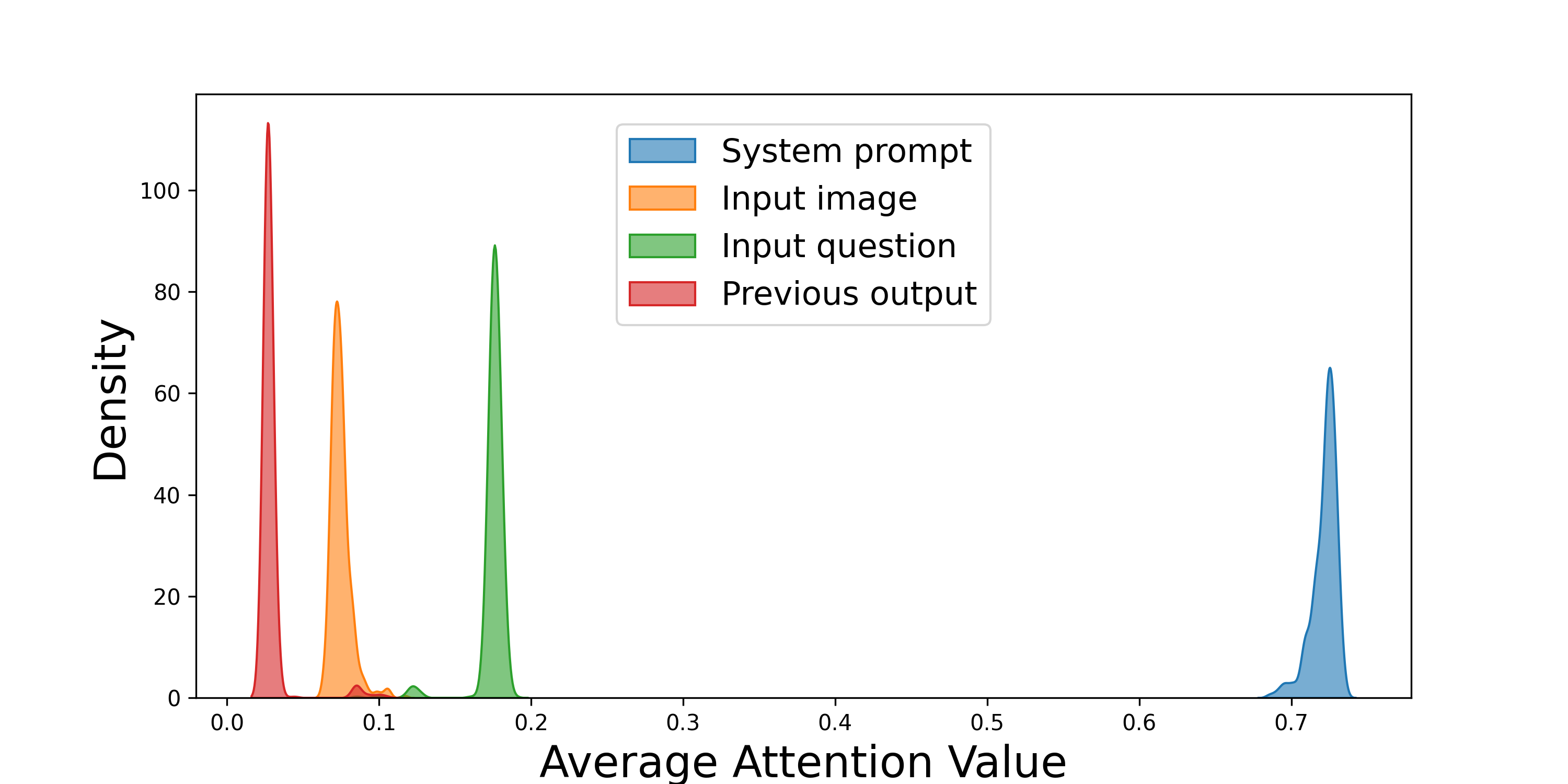}
\caption{\textbf{The overall attention allocation over different types of tokens.} We randomly select 500 images from MSCOCO \cite{lin2014mscoco} and ask the LLaVA-1.5-7B model to respond according to the images. We set the maximum number of new tokens to 20 for simplicity and collect the average attention values using Eqn.\ref{equa:average_attention}. }
\label{fig:imbalanced_att_allocation}
\vskip 0.1in
\end{figure}

As shown in Fig.\ref{fig:imbalanced_att_allocation}, it can be seen that the LVLM pays the greatest attention to the system prompt, which signifies the general target at the beginning of every task and, as such, does not contain any useful semantic information. Although it may seem counterintuitive, this pattern of attention sink, as mentioned in StreamLLM \cite{xiao2023streamllm}, is proposed to appear when redundant attention has nowhere to distribute. In LVLMs, similarly, the specific pattern may also imply the emergence of attention redundancy. 

\begin{figure}
\vskip 0.01in
\centering
\includegraphics[width=\linewidth]{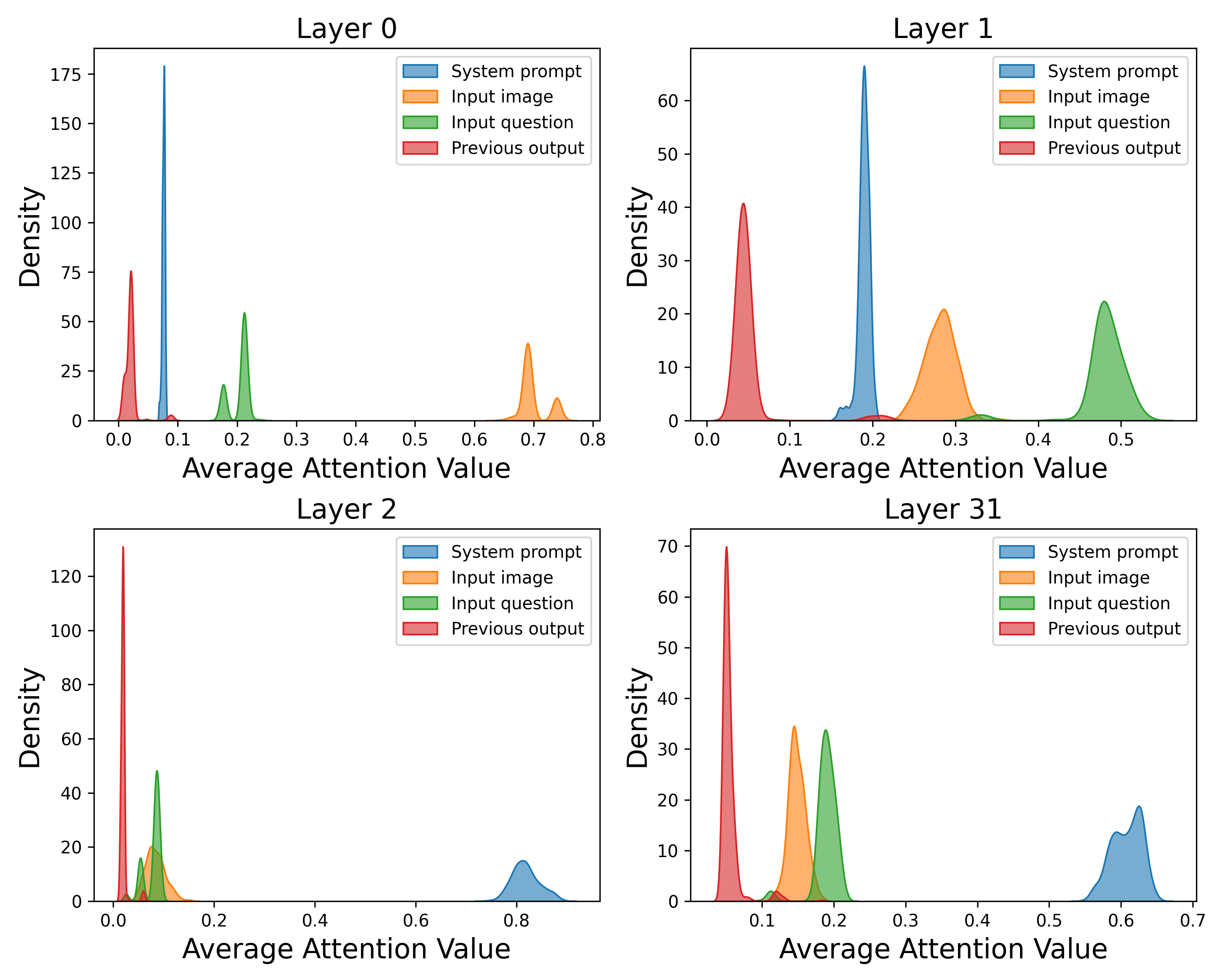}
\caption{\textbf{The attention allocation in different layers.} It can be found that the patterns in the first two layers are completely different from the ones in the 3rd and subsequent layers.}
\label{fig:layers_att_allocation}
\vskip -0.1in
\end{figure}

\textbf{Abundant Attention Patterns in shallow Layers.}
We continue to visualize the attention allocation in each layer and find that the attention sink suddenly emerges in the third layer. As depicted in Fig.\ref{fig:layers_att_allocation}, the LVLM predominantly allocates its attention to the image in the first layer, while in the second layer, the majority of attention is directed towards the input question. In other words, the LVLM completes the major semantic information understanding of the image and user's question in the first two layers, in turn. 

\begin{figure*}
\vskip 0.1in
\centering
\includegraphics[width=0.9\linewidth]{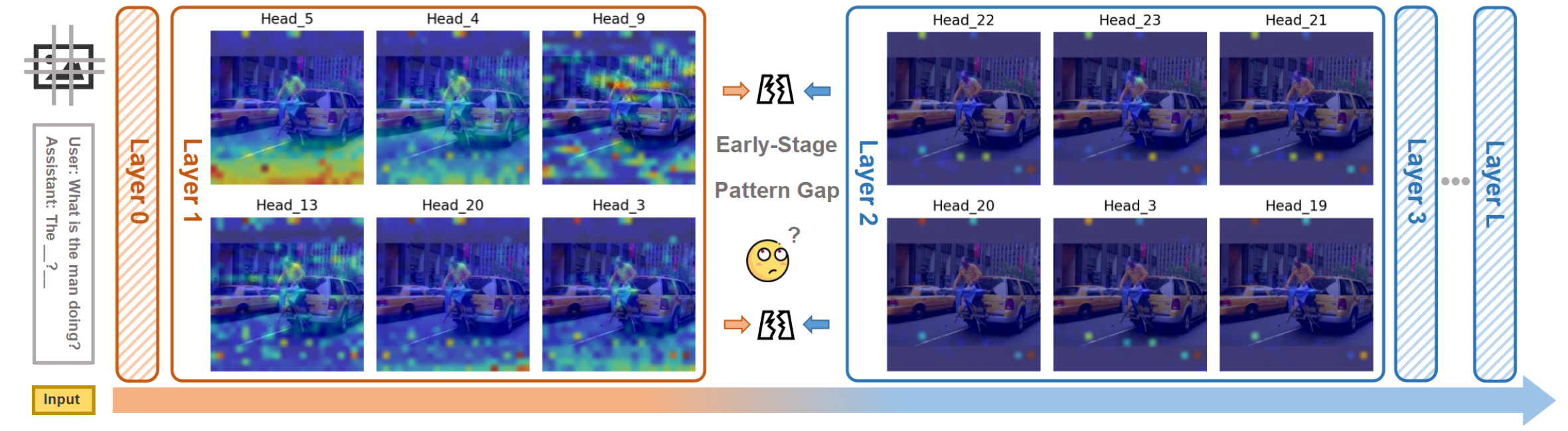}
\caption{\textbf{The attention heatmap obtained from the 2nd and the 3rd layer of the LVLM when generating the token ``[man]''.} We select the heatmap in the first six heads with the largest average attention value. It can be seen that the model's recognition of the image rapidly stabilizes, instead of being gradually formed from shallow layers to deep layers.}
\label{fig:layer_pattern_comparison}
\vskip 0.01in
\end{figure*}

To figure out how the LVLM attends to the image in each layer when conducting generation, we further visualize the attention between image patches and specific tokens. We follow the work of LVLM-Interpret \cite{ben2024lvlm-interpret}, employing an attention-based heatmap visualization technique. The implementation details of our visualization can be found in Appendix Sec.\ref{sec:appendix-visualization details}

The visualization results presented in Fig.\ref{fig:layer_pattern_comparison} show the rich attention patterns of the LVLM in the second layer, indicating that the model has developed a comprehensive understanding of the image. In contrast, the model pays nearly no attention to the image in subsequent layers, even in the third layer. However, we also find that the LVLM tends to be easily attracted to the end parts of the image token sequence in some individual heads, and even that they were meaningless padding. More visualization results are available in Appendix Sec.\ref{Sec:appendix-more visualization}.

\subsection{Focused Perception via Token Selection}
Stemming from these observations, we reasonably infer that \emph{these LVLMs' processing of images in the deep layers is possibly redundant and that not all image tokens are required during each token generation}. Actually, in most cases, only some of the key local information in the image has a decisive impact on the final response, which is also demonstrated in previous works \cite{chen2025fastv, yu2025api}. For instance, API \cite{yu2025api} enhances the original input image by overlaying a text-query-guided attention heatmap, effectively guiding LVLMs to better focus and perceive relevant information. Nevertheless, additional models are incorporated to assist in the API, which is not anticipated on our end. 

Instead, we prefer to leverage the intrinsic recognition of the LVLM to achieve focused perception. Intuitively, the abundant attention patterns mentioned in Sec.\ref{empirical analysis} might be natural guidance for selecting image tokens. As a result, we intend to follow FastV \cite{chen2025fastv}, extracting the attention map in the second layer during the generation process and simply computing the average attention score that each image token received as the criterion $\psi$:
\begin{equation}
\psi_{v_j} = \frac{1}{H} \sum_{i=1}^{H} a^{i,2}_{t,v_{j}},
\end{equation}
where $v_{j}\in X_{V}$ and $a^{i,2}_{t,v_{j}}$ represents the attention score the $t$-th token attends to the specific image token $v_{j}$ in the $i$-th head of the second layer. Subsequently, we simply sample the top-$K$ tokens based on $\psi$, which is denoted as:
\begin{equation}
 I_K = \mathop{\arg\max}\limits_{j} \psi_{v_j},   
\end{equation}
where $I_K$ represents the indices of the $K$ tokens selected. Consequently, we obtain $\Tilde{X}_{V}=\{v_j|j\in I_K \}$ and mask the LVLM's attention to those dispensable image tokens out of $\Tilde{X}_{V}$, so that the model can focus on key image regions in deep layers, minimizing certain redundancies.

\subsection{Perception Enhancement via Logit Refinement}
The dominant character of LLMs within LVLMs leads to the unavoidable occurrence of language prior, making some of the visual information neglected by LVLMs. To mitigate the negative impact, we use Contrastive Decoding \cite{li2023cd}, which aims to achieve logit refinement during the sampling process of decoding. 

The single step of LVLM's improved autoregressive token generation is formulated as follows:
\begin{equation}
\Tilde{p}_{t} = \rho \left[(1+\alpha)\mathrm{logits}_{\phi}({y}_{t}|\ast,\Tilde{X}_{V})
-\alpha\mathrm{logits}_{\phi}({y}_{t}|\ast)\right],
\end{equation}
where $\Tilde{p}_{t}$ is the calibrated probability distribution. Note that $\mathrm{logits}_{\phi}({y}_{t}|\ast) \cong \mathrm{logits}_{\phi}({y}_{t}|{y}_{< t},{X}_{I})$ is the logit distribution calculated directly with textual part and we achieve this by masking LVLM's attention to all the image tokens before the first layer. This attenuates the influence of LLM's pre-trained knowledge and thus encourages the model to focus more on the selected visual information. 

Furthermore, we introduce an adaptive plausibility constraint \cite{li2023cd} to neutralize the potential excessive intervention of our approach:
\begin{equation}
\mathcal{V}_{h}(y_{<t})=\{ y_t \in \mathcal{V}: p_{\phi}(y_t|\star)
\geq 
\beta \max_{[T]} p_{\phi}([T]|\star)\},
\end{equation}
where $p_{\phi}(y_t|\star)\cong p_{\phi}(y_t|{y}_{< t},{X}_{V},{X}_{I})$ and $p_{\phi}([T]|\star) \cong p_{\phi}([T]|{y}_{< t},{X}_{V},{X}_{I}))$. $\mathcal{V}$ is the original token vocabulary of the LVLM and $\mathcal{V}_{h}(y_{<t})$ represents the truncated one. $\beta \in[0,1]$ is a hyperparameter to control the truncation level of the probability distribution. Larger $\beta$ values indicate more aggressive truncation, retaining tokens of high probability. Therefore, the final formulation can be obtained:
\begin{equation}
y_{t} \sim \Tilde{p}_{t}~\text{~s.t.~} y_t \in \mathcal{V}_{h}(y_{<t}),
\end{equation}
Our full method dynamically enhances the model's region-aware perception, introducing a self-correction mechanism, in which non-visually centralized hallucinations are identified and mitigated via contrastive decoding. The additional integration of adaptive plausibility constraint further ensures the stability of the sampling process by taking the LVLM's confidence level into account, thereby establishing a more reliable generative paradigm.

\section{Experiments}
\subsection{Experimental Settings}
\textbf{Models.}
We choose 4 of the most popular LVLMs, namely LLaVA-1.5-7B \cite{liu2024improved-llava}, LLaVA-1.5-13B, Qwen-VL-7B \cite{bai2023qwen-vl} and MiniGPT-4-7B \cite{zhu2023minigpt}, to validate our approach. For the LLM, LLaVA-1.5 and MiniGPT4 are both built upon Vicuna, while Qwen-VL is the multimodal version of Qwen. 
For the image feature adapter, both LLaVA-1.5 models leverage MLP layers to bridge the gap between vision and language modalities, maintaining the number of image tokens (i.e., 576) before and after projection. In contrast, Qwen-VL-7B chooses a single-layer cross-attention module to compress the image features, reducing the number of image tokens to 256 for efficiency. MiniGPT-4 radically compresses the number of image tokens to 32.

\textbf{Datasets and Metrics.}
We evaluate our method on two popular binary-choice benchmarks, namely POPE \cite{li2023pope} and MME \cite{fu2023mme}. \textbf{POPE}, the polling-based object probing evaluation benchmark, aims to assess LVLMs' visual object-based perception capabilities. The entire dataset is divided into 3 splits, random, popular, and adversarial, where nonexistent objects are randomly selected, frequency-based, or frequently co-occurring with existing objects. LVLMs are required to respond to 3,000 questions asking for the existence of different objects in 500 randomly selected images for each split. \textbf{MME} contains 2,374 manually designed visual tasks involving 14 different aspects, divided into $\mathrm{MME}^P$ and $\mathrm{MME}^C$ respectively focusing on perception and cognition. It is worth noting that two questions with contradictory answers are devised for each image in MME to assess whether an LVLM indeed has correct recognition. For POPE, we use accuracy, precision, recall, and F1-score. For MME, we use accuracy+ in addition to accuracy, which requires the LVLM to answer both questions correctly in one case. 

For open-ended verification, we use CHAIR (i.e., the Caption Hallucination Assessment with Image Relevance) \cite{rohrbach2018chair} to evaluate object hallucination in image captioning tasks, including:
\begin{equation}
    \text{CHAIR}_I = \frac{|\text{\{hallucinated objects\}}|}{|\text{\{all annotated objects\}}|},
\end{equation}
\begin{equation}
    \text{CHAIR}_S = \frac{|\text{\{captions w/ hallucinated objects\}}|}{|\text{\{all captions\}}|},
\end{equation}
\begin{equation}
    \text{RECALL}_I = \frac{|\text{\{all mentioned annotated objects\}}|}{|\text{\{all annotated objects\}}|},
\end{equation}
where $\text{RECALL}_I$ is additionally considered for more comprehensive evaluation.

\textbf{Baselines.}
We compare our approach with three advanced training-free decoding methods, namely VCD \cite{leng2024vcd}, ICD \cite{wang2024ICD} and PAI \cite{liu2025pai}. VCD is the first to introduce contrastive decoding in LVLMs to mitigate hallucinations. ICD obtains the negative logits through distorted instruction. PAI additionally proposes making LVLMs pay more attention to the input image by increasing the overall attention values of the image tokens. Our approach features a similar paradigm but with more emphasis on enhancing the model's local perception.

\textbf{Implementation Details.}
We consistently set $\alpha = 1$ and $\beta = 0.1$.
Considering that different types of LVLMs have varying lengths of image tokens, we dynamically adjust the number of top-$K$ tokens by modifying a specific ratio. Specifically, we reduce 25\% image tokens in all LVLMs.

\begin{table}[htbp]
\caption{We conduct the ablation study of our method on LLaVA-1.5-7B. ``CD'' represents if ``Contrastive Decoding'' is activated and ``ACC'' means accuracy.}
\vskip 0.1in
\begin{center}
\begin{small}
\begin{sc}
\adjustbox{max width=\textwidth}{
\begin{tabular}{l c| c| c c}
\toprule
\multicolumn{2}{l}{\textbf{Model Settings}} & \textbf{POPE}  & \multicolumn{2}{c}{\textbf{MME}} \\
\cmidrule{1-5}
$K$ &CD &ACC  &ACC &ACC+ \\
\midrule
\multicolumn{1}{r}{GUESS} & & 50.00 & 1400.00 &350.00\\
576 (100\%) &$\times$ & 82.13   & 1916.50 & 658.98 \\
432 (75\%) &$\times$ & 82.38   & 1880.36 & 631.76 \\
432 (75\%)&\checkmark & \textbf{85.74}   & 1986.74 & 699.34 \\
288 (50\%)&$\times$  & 80.00   & 1875.04 & 653.94 \\ 
288 (50\%)&\checkmark & 83.88  & \textbf{2009.09} & \textbf{717.99} \\
72 (12.5\%)&$\times$  & 70.24   & 1746.37 & 563.45 \\ 
\bottomrule
\end{tabular}
}
\end{sc}
\end{small}
\end{center}
\label{tab: ablation analysis}
\end{table}

\begin{table*}[t!]
\caption{\textbf{A comparative analysis on POPE.} Our approach consistently improves model performance in object perception compared to all original models. Note that we do not include ICD in Qwen-VL due to implementation issues with the model architecture.}
\label{tab:pope}
\vskip 0.1in
\begin{center}
\begin{small}
\begin{sc}
\resizebox{0.952\textwidth}{!}{
\begin{tabular}{lccccccc}
\toprule
\textbf{Datasets} & \textbf{POPE} & \textbf{Models} &\textbf{Decoding Methods} & \textbf{Accuracy~(\%)} & \textbf{Precision~(\%)} & \textbf{Recall~(\%)} & \textbf{F1 Score~(\%)} \\
\midrule
\multirow{50}{*}{MSCOCO} 
& \multirow{16}{*}{Random} 
& \multirow{4}{*}{LLaVA-1.5-7B} 
& Vanilla & {84.33} & {90.62} & 76.6 & {83.02} \\
& & &ICD & 85.53 & 94.12 & 75.80 & 83.97 \\
& & &VCD & {86.63} & {92.24} & \textbf{80.00} & {85.69} \\
& & &\textbf{{\scshape MINT}(Ours)} & \cellcolor{mycolor}\textbf{87.60} & \cellcolor{mycolor}\textbf{96.23} & \cellcolor{mycolor}78.27 & \cellcolor{mycolor}\textbf{86.32} \\
\cmidrule{4-8}
& & \multirow{4}{*}{LLaVA-1.5-13B} 
& Vanilla & 83.63 & 86.01 & 80.33 & 83.07 \\
& & &ICD & 84.80 & 86.20 & \textbf{82.87} & 84.50 \\
& & &VCD & 85.57 & 87.86 & 82.53 & 85.12 \\
& & &\textbf{{\scshape MINT}(Ours)} & \cellcolor{mycolor}\textbf{87.23} & \cellcolor{mycolor}\textbf{95.89} & \cellcolor{mycolor}77.80 & \cellcolor{mycolor}\textbf{85.90} \\
\cmidrule{4-8}
& & \multirow{3}{*}{Qwen-VL} 
& Vanilla & 84.87 & 95.16 & 73.47 & 82.92\\
& & & VCD & \textbf{86.03} & 95.46 & \textbf{75.67} & \textbf{84.42}\\
& & &\textbf{{\scshape MINT}(Ours)} & \cellcolor{mycolor}85.17 & \cellcolor{mycolor}\textbf{97.14} &\cellcolor{mycolor}72.47 &\cellcolor{mycolor}83.01 \\
\cmidrule{4-8}
& &\multirow{4}{*}{MiniGPT-4}
& Vanilla & 58.03 & 64.71 & 35.33 & 45.71\\
& & & ICD & 61.03 & 61.73 & 58.07 & 59.84\\
& & & VCD & \textbf{62.17} & 62.46 & \textbf{61.00} & \textbf{61.72}\\
& & &\textbf{{\scshape MINT}(Ours)} & \cellcolor{mycolor}61.80 & \cellcolor{mycolor}\textbf{67.25} &\cellcolor{mycolor}46.00 &\cellcolor{mycolor}54.63 \\

\cmidrule{3-8}
& \multirow{16}{*}{Popular} 
& \multirow{4}{*}{LLaVA-1.5-7B} 
& Vanilla & {82.63} & {87.74} & 75.87 & {81.37} \\
& & & ICD & 84.50 & 92.24 & 75.33 & 82.94 \\
& & & VCD & {83.93} & {86.99} & \textbf{79.80} & {83.24} \\
& & &\textbf{{\scshape MINT}(Ours)} & \cellcolor{mycolor}\textbf{86.40} & \cellcolor{mycolor}\textbf{93.47} & \cellcolor{mycolor}78.27 & \cellcolor{mycolor}\textbf{85.20} \\
\cmidrule{4-8}
& & \multirow{4}{*}{LLaVA-1.5-13B} 
& Vanilla & 83.30 & 85.05 & 80.80 & 82.87 \\
& & &ICD & 84.17 & 85.13 & 82.80 & 83.95 \\
& & &VCD & 85.00 & 86.56 & \textbf{82.87} & 84.67 \\
& & &\textbf{{\scshape MINT}(Ours)} & \cellcolor{mycolor}\textbf{86.00} & \cellcolor{mycolor}93.41 & \cellcolor{mycolor}77.47 &\cellcolor{mycolor} \textbf{84.69} \\
\cmidrule{4-8}
& &\multirow{3}{*}{Qwen-VL}
& Vanilla & 84.30 & 95.25 & 72.20 & 82.14\\
& & & VCD & \textbf{85.70} & 94.37 & \textbf{75.93} & \textbf{84.15}\\
& & &\textbf{{\scshape MINT}(Ours)} & \cellcolor{mycolor}84.83 & \cellcolor{mycolor}\textbf{95.71} &\cellcolor{mycolor}72.93 &\cellcolor{mycolor}82.78 \\
\cmidrule{4-8}
& &\multirow{4}{*}{MiniGPT-4}
& Vanilla & 56.77 & 60.95 & 37.67 & 46.56\\
& & & ICD & 58.60 & 58.41 & 59.73 & 59.06\\
& & & VCD & 59.80 & 59.42 & \textbf{61.80} & \textbf{60.59}\\
& & &\textbf{{\scshape MINT}(Ours)} & \cellcolor{mycolor}\textbf{60.00}& \cellcolor{mycolor}\textbf{63.91} &\cellcolor{mycolor}45.93 &\cellcolor{mycolor}53.45 \\

\cmidrule{3-8}
& \multirow{16}{*}{Adversarial} 
& \multirow{4}{*}{LLaVA-1.5-7B} 
& Vanilla & {79.43} & {81.25} & 76.53 & {78.82} \\
& & & ICD & 82.90 & \textbf{87.82} & 76.40 & 81.71 \\
& & & VCD & {81.07} & {81.96} & \textbf{79.67} & {80.80} \\
& & &\textbf{{\scshape MINT}(Ours)} & \cellcolor{mycolor}\textbf{83.23} & \cellcolor{mycolor}87.40 & \cellcolor{mycolor}77.67 & \cellcolor{mycolor}\textbf{82.24} \\
\cmidrule{4-8}
& & \multirow{4}{*}{LLaVA-1.5-13B} 
& Vanilla & 79.77 & 79.20 & 80.73 & 79.96 \\
& & &ICD & 81.33 & 80.32 & \textbf{83.00} & 81.64 \\
& & &VCD & 81.13 & 80.01 & 83.00 & 81.48 \\
& & & \textbf{{\scshape MINT}(Ours)} & \cellcolor{mycolor}\textbf{84.03} & \cellcolor{mycolor}\textbf{88.82} & \cellcolor{mycolor}77.87 & \cellcolor{mycolor} \textbf{82.98} \\
\cmidrule{4-8}
& &\multirow{3}{*}{Qwen-VL}
& Vanilla & 82.50 & 89.67 & 73.47 & 80.77\\
& & & VCD & \textbf{83.53} & 89.17 & \textbf{76.33} & \textbf{82.25}\\
& & &\textbf{{\scshape MINT}(Ours)} & \cellcolor{mycolor}82.57 & \cellcolor{mycolor}\textbf{91.65} &\cellcolor{mycolor}71.67 &\cellcolor{mycolor}80.44 \\
\cmidrule{4-8}
& &\multirow{4}{*}{MiniGPT-4}
& Vanilla & 54.87 & 57.92 & 35.60 & 44.10\\
& & &ICD & 57.70 & 57.43 & 59.53 & 58.46 \\
& & & VCD & 57.70 & 57.19 & \textbf{61.27} & \textbf{59.16}\\
& & &\textbf{{\scshape MINT}(Ours)} & \cellcolor{mycolor}\textbf{58.60} & \cellcolor{mycolor}\textbf{61.62} &\cellcolor{mycolor}45.60 &\cellcolor{mycolor}52.41 \\
\bottomrule
\end{tabular}
}
\end{sc}
\end{small}
\end{center}
\vskip -0.1in
\end{table*}

\subsection{Experimental Results}
\textbf{Half Tokens Can Be Enough or Better.} 
We progressively reduce the number of image tokens available to the model, as presented in Tab.\ref{tab: ablation analysis}. The experimental results show that the model's performance on the datasets only experiences a slight decline, or even an improvement, despite halving the number of tokens. This suggests that there is indeed token redundancy during the generation process. Meanwhile, the token selection module does effectively direct the model’s focus to the most relevant regions. Furthermore, the model demonstrates enhanced performance when encouraged to focus more on the selected image tokens, following the implementation of contrastive decoding. This result further substantiates the efficacy of the simple token selection module. However, it is also important to note that the model's performance significantly deteriorates when the number of image tokens is reduced beyond a certain threshold. This indicates that, in some instances, critical image tokens may have been erroneously filtered out, highlighting that there is still room for improvement in the token selection process.

\textbf{Superior Performance on Object Perception.}
Compared to vanilla models, our approach significantly enhances their ability to perceive object existence, as demonstrated in Tab.\ref{tab:pope}. Notably, our method achieves an accuracy improvement of nearly 4\% on the LLaVA-1.5 model across both parameter sizes. Furthermore, when compared to either the ICD or the VCD method, our approach still demonstrates an obvious improvement even though some visual tokens are discarded. Additionally, we implement our method to Qwen-VL, an LVLM with a fundamentally different architecture, yielding a modest improvement over the original model and comparable performance to the VCD method. Considering the architectural differences, one possible explanation is that the attention in this LVLM features a different pattern with the LLaMA models. We continue to evaluate another LLaMA-based model (i.e., MiniGPT-4), finding that our method significantly enhances its perception capability even if we reduce 25\% of its 32 image tokens. This also indicates that the phenomenon of token redundancy still occurs in such extreme compression of visual information. However, the overall poor performance also suggests that the feature alignment is not adequate and lacks effective visual compression.
\begin{figure}[t]
\vskip 0.1in
\centering
\includegraphics[width=0.95\linewidth]{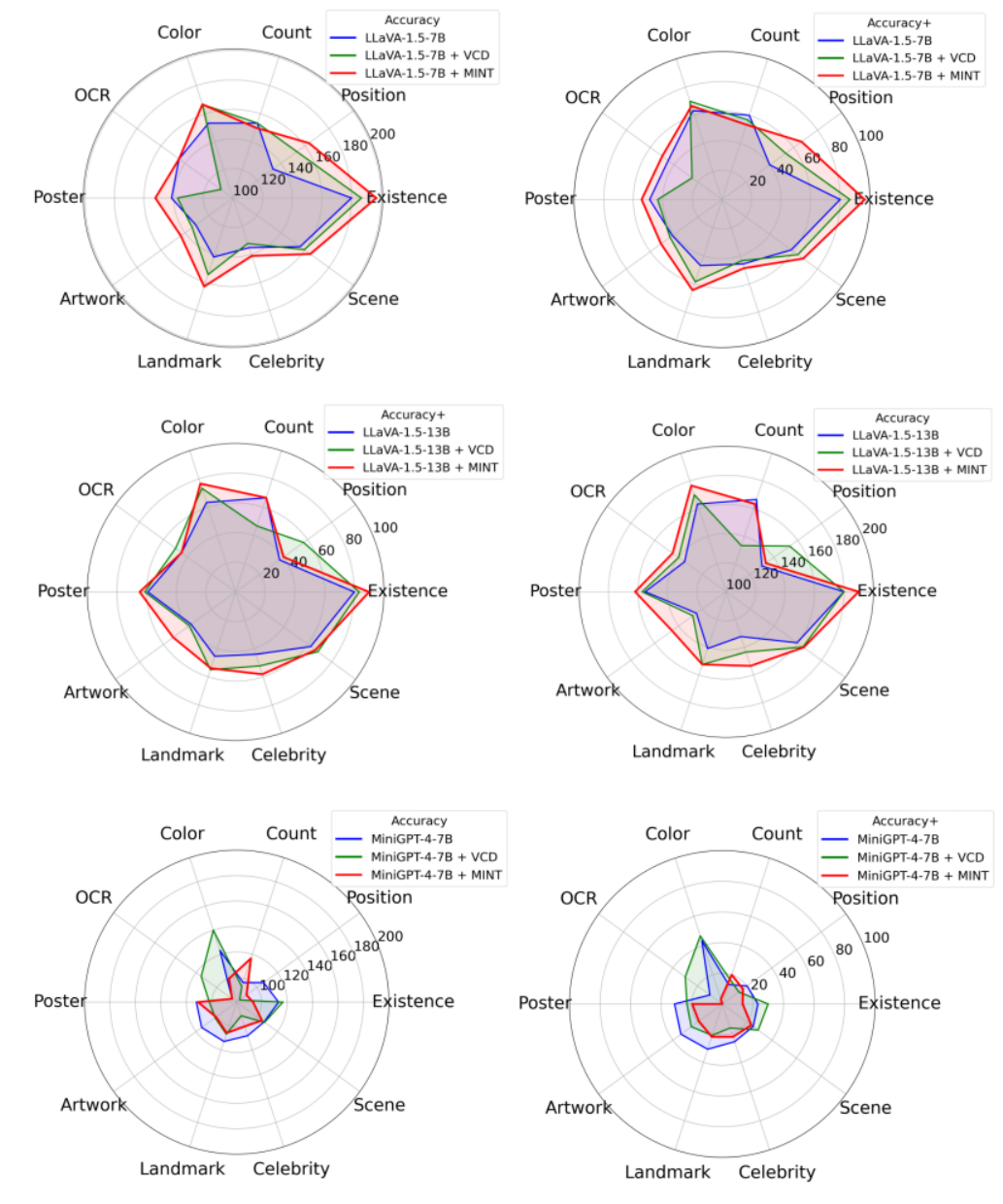}
\caption{A comparative evaluation result on $\mathrm{MME}^P$.}
\label{fig:mme_comparison}
\vskip -0.1in
\end{figure}

\begin{table}[h]
\caption{A comparative analysis of different decoding methods with LLaVA-1.5-7B on open-ended image captioning tasks.}
\vskip 0.1in
\begin{center}
\begin{small}
\begin{sc}
\resizebox{0.47\textwidth}{!}{
\begin{tabular}{c c c c}
\toprule
\textbf{Methods} & $\mathrm{\textbf{CHAIR}}_{\textbf{I}} \downarrow$  & $\mathrm{\textbf{CHAIR}}_{\textbf{S}} \downarrow$ & $\mathrm{\textbf{RECALL}}_{\textbf{I}} \uparrow$\\
\midrule
\rowcolor{mycolor1}\multicolumn{4}{c}{\textit{64 new tokens}} \\
Vanilla & 7.9 & 25.0 & 54.9 \\
ICD & 8.8 & 24.0 & 58.0 \\
VCD & 8.3 & 25.0 & 57.0 \\
PAI & 6.8 & 22.0 & 55.8 \\
\textbf{MINT(Ours)} & \textbf{4.7} & \textbf{18.0} & \textbf{60.1} \\
\midrule
\rowcolor{mycolor1}\multicolumn{4}{c}{\textit{512 new tokens}} \\
Vanilla & 16.1 & 64.0 & 70.9 \\
ICD & 17.6 & 57.0 & 71.4 \\
VCD & 17.2 & 69.0 & 74.0 \\
PAI & 11.7 & \textbf{46.0} & 67.3 \\
\textbf{MINT(Ours)} & 14.2 & 61.0 & \textbf{77.1} \\
\textbf{MINT(Ours, K=288)} & \textbf{11.3} & 51.0 & 72.1 \\
\bottomrule
\end{tabular}}
\end{sc}
\end{small}
\end{center}
\label{tab:chair_method_comparison}
\vskip -0.1in
\end{table}

\textbf{Perception Enhancement in Various Domains.} The extra evaluation result on MME, as depicted in Fig.\ref{fig:mme_comparison}, shows that our approach surpasses the LLaVA-1.5 models and the baseline method VCD in nearly all ten subsets in $\mathrm{MME}^P$. Meanwhile, it maintains superior performance when upgrading the level of evaluation with extra consideration of consistency. However, for MiniGPT-4, it is found that our method causes an overall decline in the model's performance. We infer that the model's original poor comprehension in these domains causes ineffective information compression and incorrect token selection, thus accidentally excluding some key visual information. In contrast, when sufficient tokens containing rich semantic information are available for selection, our approach dynamically amplifies key visual points, enabling the model to make more informed decisions.

\textbf{Excellence in Open-ended Captioning.}
We conduct the CHAIR evaluation on the MSCOCO dataset with rich annotations, randomly selecting 100 images and using the prompt ``Please describe the image in detail.'' to obtain the captions. Note that we evaluate the performance of the model to generate short and long captions, respectively, by controlling the maximum number of generated tokens. Specific cases are offered in Appendix Sec.~\ref{sec: case study}

\begin{table}[h!]
\caption{\textbf{A comparative analysis of different models on open-ended image captioning tasks.} It is worth noting that Qwen-VL is excluded from our evaluation, as it consistently produces extremely short captions, regardless of our instructions.}
\vskip 0.1in
\begin{center}
\begin{small}
\begin{sc}
\resizebox{0.47\textwidth}{!}{
\begin{tabular}{c c c c}
\toprule
\textbf{Methods }& $\mathrm{\textbf{CHAIR}}_{\textbf{I}} \downarrow$  & $\mathrm{\textbf{CHAIR}}_{\textbf{S}} \downarrow$ & $\mathrm{\textbf{RECALL}}_{\textbf{I}} \uparrow$ \\
\midrule
\rowcolor{mycolor1}\multicolumn{4}{c}{\textit{LLaVA-1.5-13B}} \\
Vanilla & \textbf{5.9} & \textbf{22.0} & 56.6 \\
MINT(Ours) & 6.6 & 23.0 & \textbf{57.5} \\
\midrule
\rowcolor{mycolor1}\multicolumn{4}{c}{\textit{MiniGPT-4-7B}} \\
Vanilla & 8.7 & 25.0 & 49.6 \\
MINT(Ours) & \textbf{5.3} & \textbf{17.0} & \textbf{51.6} \\
\bottomrule
\end{tabular}
}
\end{sc}
\end{small}
\end{center}
\label{tab:chair_model_comparison}
\vskip -0.1in
\end{table}

As shown in Tab.\ref{tab:chair_method_comparison}, while both the ICD and VCD methods enhance the model's visual perception ability, they do not effectively reduce hallucinations. On the other hand, the PAI method significantly mitigates hallucinations, but this comes at the cost of sacrificing the model's perceptual capabilities. In contrast, our approach outperforms all other baseline methods in short caption generation tasks, effectively reducing hallucinations while capturing more visual details. For longer caption generation tasks, our method continues to demonstrate superior performance. When the number of image tokens is reduced to 288, the model’s hallucinations are noticeably alleviated further, while its perceptual ability remains comparable to, or even better than, other methods. 

In addition, we assess the effectiveness of our approach on open-ended image captioning when applied to other LVLMs, as displayed in Tab.\ref{tab:chair_model_comparison}. Note that MiniGPT-4 shows a stronger ability to conduct open-ended generation compared to its poor performance in evaluation on benchmarks. The positive influence on MiniGPT-4-7B demonstrates that our method can further unlock the models' potential. We also note that our method brings about less benefit when applied to LLaVA-1.5-13B in this task, indicating that the model with a large size is more confident with its decoding.



\section{Conclusion}
In this paper, we first analyze the imbalanced attention allocation and attention redundancy in LVLMs. We observe that LVLMs allocate the most attention to image tokens in shallow layers, leading to strong recognition of the image itself in the early stage. Based on this observation, we propose MINT, a novel training-free method that mitigates hallucinations without relying on external tools. MINT guides the LVLM to focus on key visual elements through token selection, while also pushing the model to further concentrate on visual information using contrastive decoding. Extensive experiments across various benchmarks and LVLMs demonstrate the effectiveness of our approach. Specifically, MINT helps mitigate hallucinations in LVLMs while enabling the models to capture more details yet with fewer image tokens.

\nocite{langley00}



\bibliography{example_paper}
\bibliographystyle{icml2025}

\appendix
\onecolumn
\section{More Implementation Details.}
\subsection{Visualization.} \label{sec:appendix-visualization details}
Specifically, we first obtain the attention value sequence $a^{i,j}_{t,img}$ and resize it to a square matrix $A\in \mathbb{R}^{\sqrt{n}\times \sqrt{n}}$. The attention matrix $A$ is then normalized to the range $[0,1]$ to ensure consistent scaling across different values and converted into a heatmap using a colormap function:
\begin{equation}
    A^{\prime} =\left( A-\min(A))/(\max(A)-\min(A)\right),
\end{equation}
\begin{equation}
    H = f_{cmap}(A^{\prime}),
\end{equation}
where $H\in \mathbb{R}^{\sqrt{n} \times \sqrt{n} \times 3}$ is the resulting RGB heatmap. Subsequently, we resize $H$ to match the spatial dimension of the formatted image (e.g., $336 \times 336$) and concurrently use bicubic interpolation to preserve visual fidelity. Ultimately, $H$ is overlaid on the formatted input image with the additional alpha channel set to 50\% opacity, serving as a mask. The final visualization highlights the attention intensities over every patch of the image, providing insights into the model's focus during the token generation process. 

\subsection{Methods Settings.}
The baseline PAI is based on the Beam Search sampling strategy. For the baseline ICD and VCD, we employ the same direct sampling strategy as ours. Throughout the entire experiment, our experimental hyperparameters remain consistent. The hyperparameters are listed in the tables below:
\begin{table}[htbp]
\centering
\begin{tabular}{ll}
\toprule
\textbf{Hyperparameters} & Value \\
\midrule
$\alpha$ for Contrastive Decoding  & 1.0 \\
$\beta$ for Adaptive Plausibility Threshold & 0.1 \\
Distorted Prompt & ``You are a confused object detector.'' \\
\bottomrule
\end{tabular}
\caption{ICD settings.}
\label{tab:icd_settings}
\end{table}

\begin{table}[htbp]
\centering
\begin{tabular}{ll}
\toprule
\textbf{Hyperparameters} & Value \\
\midrule
$\alpha$ for Contrastive Decoding  & 1.0 \\
$\beta$ for Adaptive Plausibility Threshold & 0.1 \\
Diffusion Noise Step & 500 \\
\bottomrule
\end{tabular}
\caption{VCD settings.}
\label{tab:vcd_settings}
\end{table}

\begin{table}[htbp]
\centering
\begin{tabular}{ll}
\toprule
\textbf{Hyperparameters} & Value \\
\midrule
$\lambda$ for Classifier-Free Guidance  & 1.1 \\
$\alpha$ for attention amplification & 0.2 \\
Start layer for attention amplification & 2 \\
End layer for attention amplification & 32 \\
Number of beams & 5 \\
\bottomrule
\end{tabular}
\caption{PAI Settings.}
\label{tab:pai_settings}
\end{table}

\section{More Visualization Results in Empirical Analysis.} \label{Sec:appendix-more visualization}
\begin{figure}[htbp]
\centering
\includegraphics[width=0.83\linewidth]{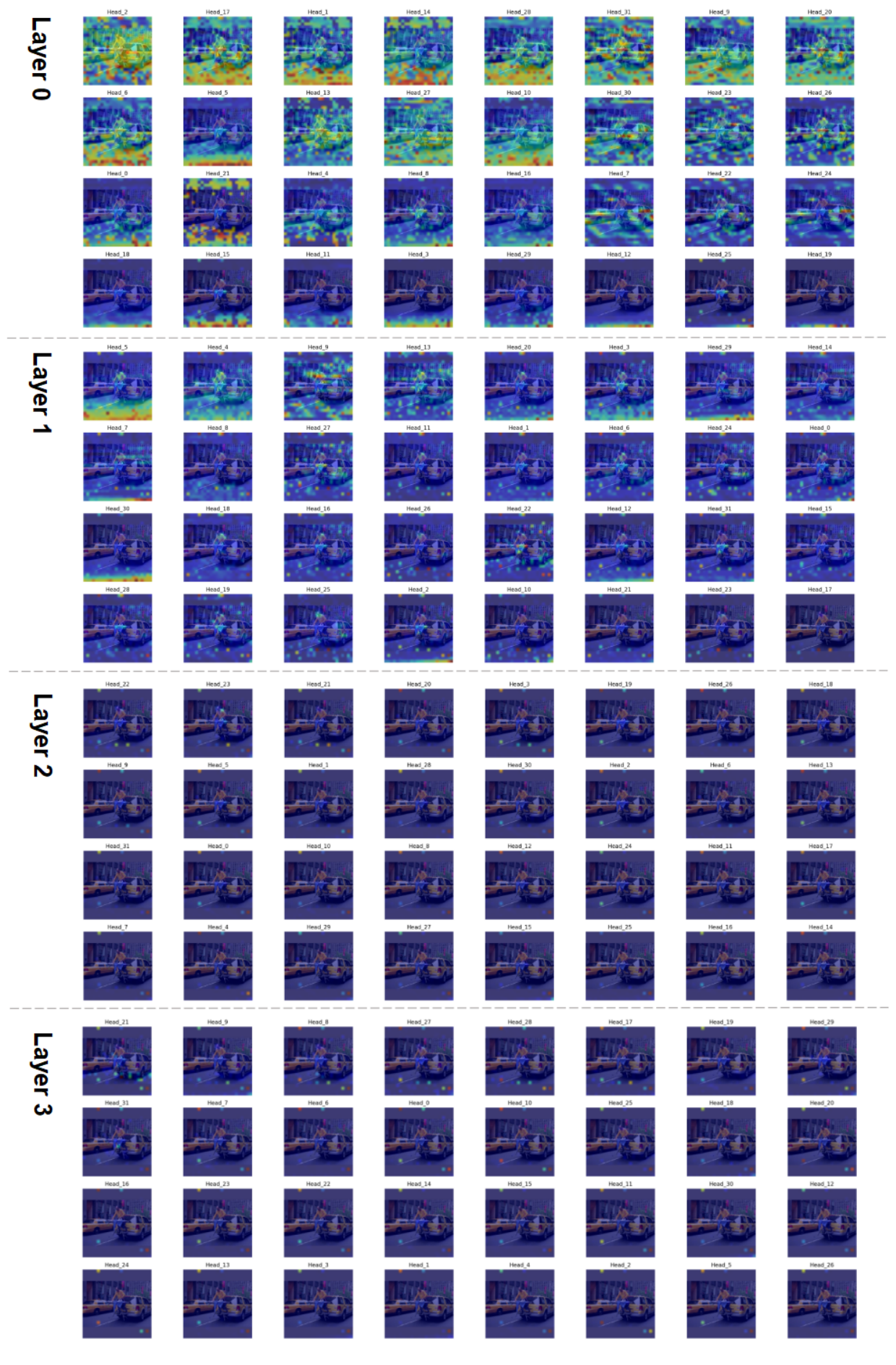}
\caption{More visualization results of query-to-patch attention map in shallow layers.}
\label{fig:visualization_results_shallow}
\end{figure}

\begin{figure*}[htbp]
\centering
\includegraphics[width=0.83\linewidth]{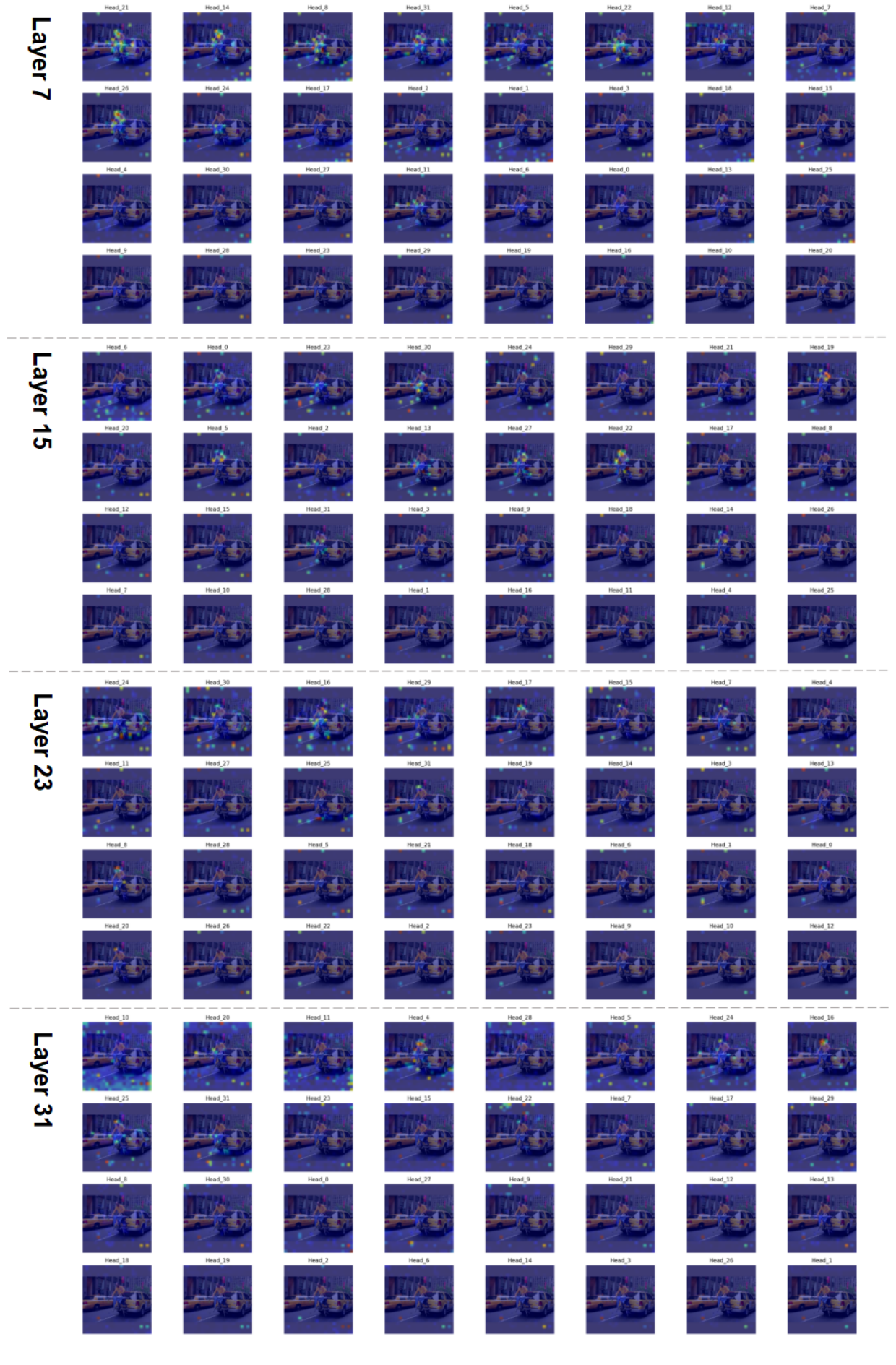}
\caption{More visualization results of query-to-patch attention map in deep layers.}
\label{fig:visualization_results_deep}
\end{figure*}

The detailed results of visualization in our empirical analysis are shown in Fig.\ref{fig:visualization_results_shallow} and Fig.\ref{fig:visualization_results_deep}. We can see that the model's attention patterns on the image undergo a rapid transition within the first four layers, shifting from being highly abundant to nearly negligible. Even in the deeper layers, the model only makes subtle adjustments to its attention, which are far less substantial compared to the rich perception observed in the initial two layers. Interestingly, we observe that different attention heads exhibit their own unique patterns. However, this requires further validation through additional statistical observations on a large number of cases in future studies.

\section{Ablation Study Details.}
The detailed results of our ablation study on POPE are shown in Tab.\ref{tab:detailed_ablation_study_results}. Note that we also perform ablation experiments of our method on Qwen-VL, a model with a completely different architecture, and similarly find that it exhibits a certain degree of token redundancy.

\begin{table}[htbp]
\caption{The detailed experimental results of ablation study on POPE.}
\begin{center}
\begin{small}
\begin{sc}
\adjustbox{max width=\textwidth}{
\begin{tabular}{l c c| c c |c c| c c}
\toprule
\multirow{3.5}{*}{Models} &\multicolumn{2}{c}{Settings} & \multicolumn{6}{c}{POPE} \\
\cmidrule{2-9}
&\multirow{2.5}{*}{$K$} &\multirow{2.5}{*}{CD} &\multicolumn{2}{c}{Random} & \multicolumn{2}{c}{Popular} &\multicolumn{2}{c}{Adversarial} \\
\cmidrule{4-9}
& & & ACC &F1-score & ACC &F1-score & ACC &F1-score \\
\midrule
LLaVA-1.5-7B &576 &$\times$ & 84.33 &83.02 &82.63 & 81.37 &79.43 &78.82 \\
LLaVA-1.5-7B &432 &$\times$ & 84.17 &82.57 &82.77 & 81.18 &80.20 &79.27 \\ 
LLaVA-1.5-7B &432 &\checkmark  & 87.60 &86.32 &86.40 & 85.20 &83.23 &82.24 \\
LLaVA-1.5-7B &288 &$\times$ & 80.87 &78.34 &81.43 & 79.19 &77.80 &75.69 \\ 
LLaVA-1.5-7B &288 &\checkmark  & 84.83 &82.73 &84.57 & 82.50 &82.23 &80.43 \\
LLaVA-1.5-7B &72 &$\times$ & 69.37 &58.47 &71.27 & 61.52 &70.07 &60.75 \\ 
\midrule
Qwen-VL-7B &256 &$\times$ & 84.87 &82.92 &84.30 & 82.14 &82.50 &80.77 \\
Qwen-VL-7B &224 &$\times$ & 84.20 &81.95 &84.07 & 81.85 &82.17 &80.04 \\
Qwen-VL-7B &192 &$\times$ & 83.13 &80.39 &83.50 & 81.12 &81.83 &79.41 \\
Qwen-VL-7B &128 &$\times$ & 80.77 &76.87 &80.67 & 76.76 &79.77 &76.07 \\
Qwen-VL-7B &64 &$\times$ &74.73  &67.13 &74.93 & 67.25 &73.93 &66.44 \\
\bottomrule
\end{tabular}
}
\end{sc}
\end{small}
\end{center}
\label{tab:detailed_ablation_study_results}
\end{table}

\section{Detailed Results on MME.}
The detailed results of the comparative analysis on $\mathrm{MME}^P$ are shown in Tab.\ref{tab:mme_detailed_results_acc} and Tab.\ref{tab:mme_detailed_results_acc+}. This benchmark is manually designed without using public datasets to avoid data leakage. We additionally use this multimodal benchmark to more comprehensively evaluate how our method influences other perceptual abilities beyond object perception.

\begin{table}[htbp]
\centering
\caption{Detailed results of evaluation using accuracy on $\mathrm{MME}^P$.}
\adjustbox{max width=0.95\textwidth}{
\begin{tabular}{l c | c c c c c c c c c c}
\toprule
Models & Methods & Existence &Position &Count &Color &OCR &Poster &Artwork &Landmark &Celebrity &Scene \\
\midrule
\multirow{3}{*}{LLaVA-1.5-7B} & Vanilla &180.00 &133.34 &153.33 &153.33 &145.00 &141.49 &131.00 &142.00 &135.29 &156.00 \\
 & VCD &186.67 &153.34 &153.33 &166.66 &110.00 &137.42 &134.00 &154.50 &132.35 &159.50\\
 & MINT (Ours) &196.67 &163.34 &150.00 &166.67 &145.00 &152.38 &143.50 &163.00 &141.18 &164.50\\
\cmidrule{2-12}
\multirow{3}{*}{LLaVA-1.5-13B} & Vanilla &180.00 &130.00 &166.67 &163.33 &135.00 &155.11 &125.00 &141.00 &132.35 &159.50\\
 & VCD &180.00 &153.33 &133.33 &170.00 &140.00 &157.14 &128.00 &152.50 &143.53 &164.50\\
 & MINT (Ours) &190.00 &133.34 &163.33 &176.67 &145.00 &161.91 &145.50 &152.50 &153.53 &165.00\\
 \cmidrule{2-12}
\multirow{3}{*}{MiniGPT-4-7B} & Vanilla &113.34 &106.66 &96.67 &123.33 &85.00 &112.24 &114.50 &113.00 &108.23 &107.00 \\
 & VCD &116.67, &83.33, &93.34, &140.00, &115.00, &102.04, &101.50, &107.00, &91.77, &107.50\\
 & MINT (Ours) &93.33 &90.00 &116.66 &100.00 &85.00 &110.88 &100.50 &106.00 &100.59 &105.00\\
\bottomrule
\end{tabular}}
\label{tab:mme_detailed_results_acc}
\end{table}

\begin{table}[htbp]
\centering
\caption{Detailed results of evaluation using accuracy+ on $\mathrm{MME}^P$.}
\adjustbox{max width=0.95\textwidth}{
\begin{tabular}{l c | c c c c c c c c c c}
\toprule
Models & Methods & Existence &Position &Count &Color &OCR &Poster &Artwork &Landmark &Celebrity &Scene \\
\midrule
\multirow{3}{*}{LLaVA-1.5-7B} & Vanilla &80.00 &40.00 &60.00 &63.33 &45.00 &48.98 &41.50 &47.00 &45.88 &58.00 \\
 & VCD &86.67 &53.33 &56.67 &70.00 &25.00 &43.54 &43.50 &58.50 &43.53 &63.50\\
 & MINT (Ours) &96.67 &66.67 &53.33 &66.67 &50.00 &54.42 &51.00 &64.50 &48.82 &68.00\\
\cmidrule{2-12}
\multirow{3}{*}{LLaVA-1.5-13B} & Vanilla &80.00 &36.67 &66.67 &63.33 &45.00 &59.18 &37.00 &45.50 &44.12 &62.50 \\
 & VCD &83.33 &56.67 &46.67 &73.33 &50.00 &61.22 &38.50 &55.00 &52.35 &68.50\\
 & MINT (Ours) &90.00 &40.00 &66.67 &76.67 &45.00 &64.63 &52.00 &54.00 &58.24 &66.50\\
 \cmidrule{2-12}
\multirow{3}{*}{MiniGPT-4-7B} & Vanilla &23.33 &20.00 &13.33 &43.33 &10.00 &31.29 &33.50 &31.00 &25.88 &25.00\\
 & VCD &30.00 &13.33 &16.67 &46.67 &30.00 &23.13 &25.15 &21.50 &16.47 &29.00\\
 & MINT (Ours) &13.33 &16.67 &20.00 &3.33 &0.00 &19.73 &19.00 &22.50 &22.50 &23.50\\
\bottomrule
\end{tabular}}
\label{tab:mme_detailed_results_acc+}
\end{table}

\section{Case Study} \label{sec: case study}
\begin{figure*}[htbp]
\centering
\includegraphics[width=0.8\linewidth]{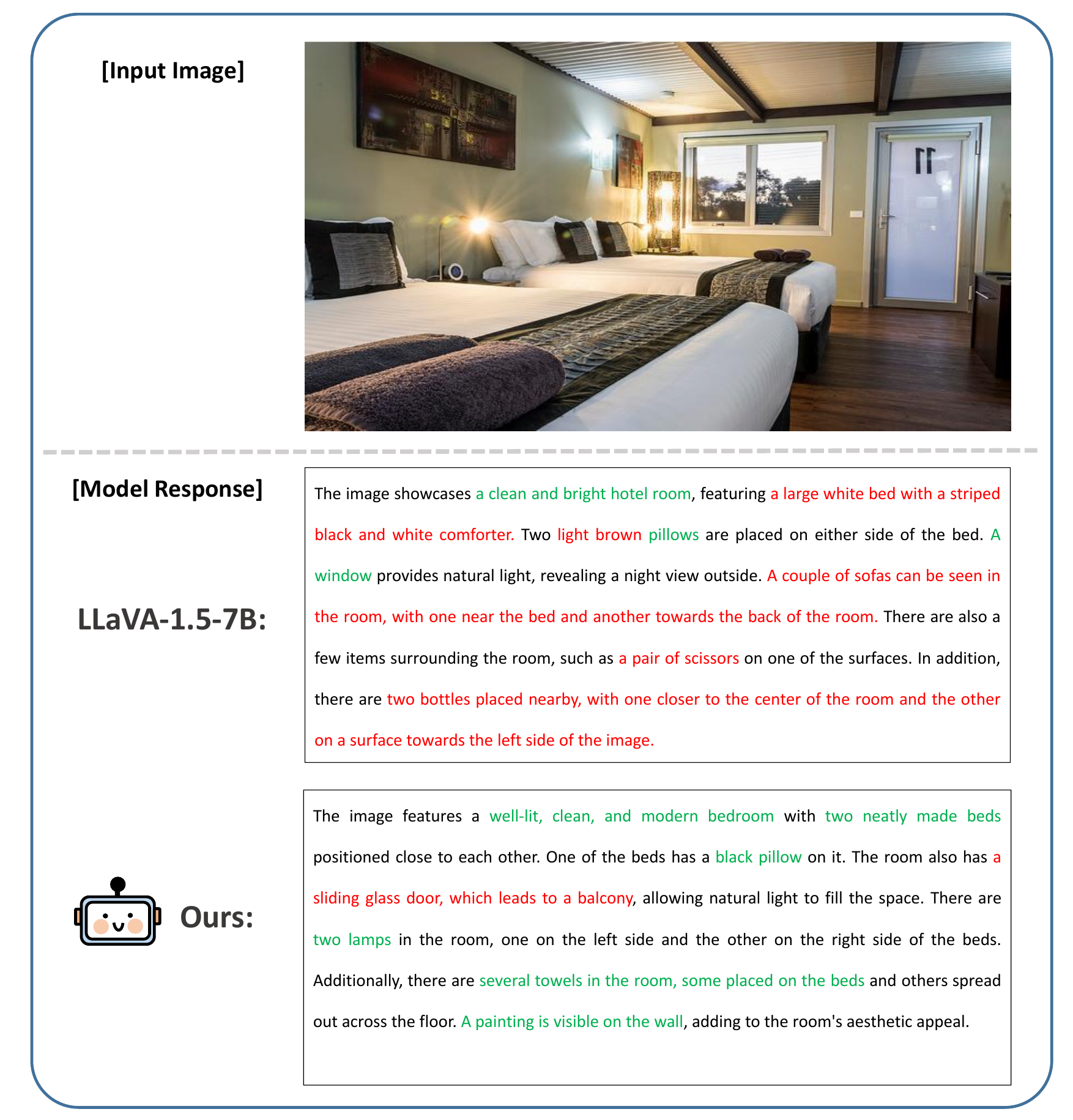}
\caption{One of the cases illustrates the effectiveness of our method. We use the prompt ``Please describe the image in detail.'' to get the caption.}
\label{fig:case_study1}
\end{figure*}

\begin{figure*}[t]
\centering
\includegraphics[width=0.8\linewidth]{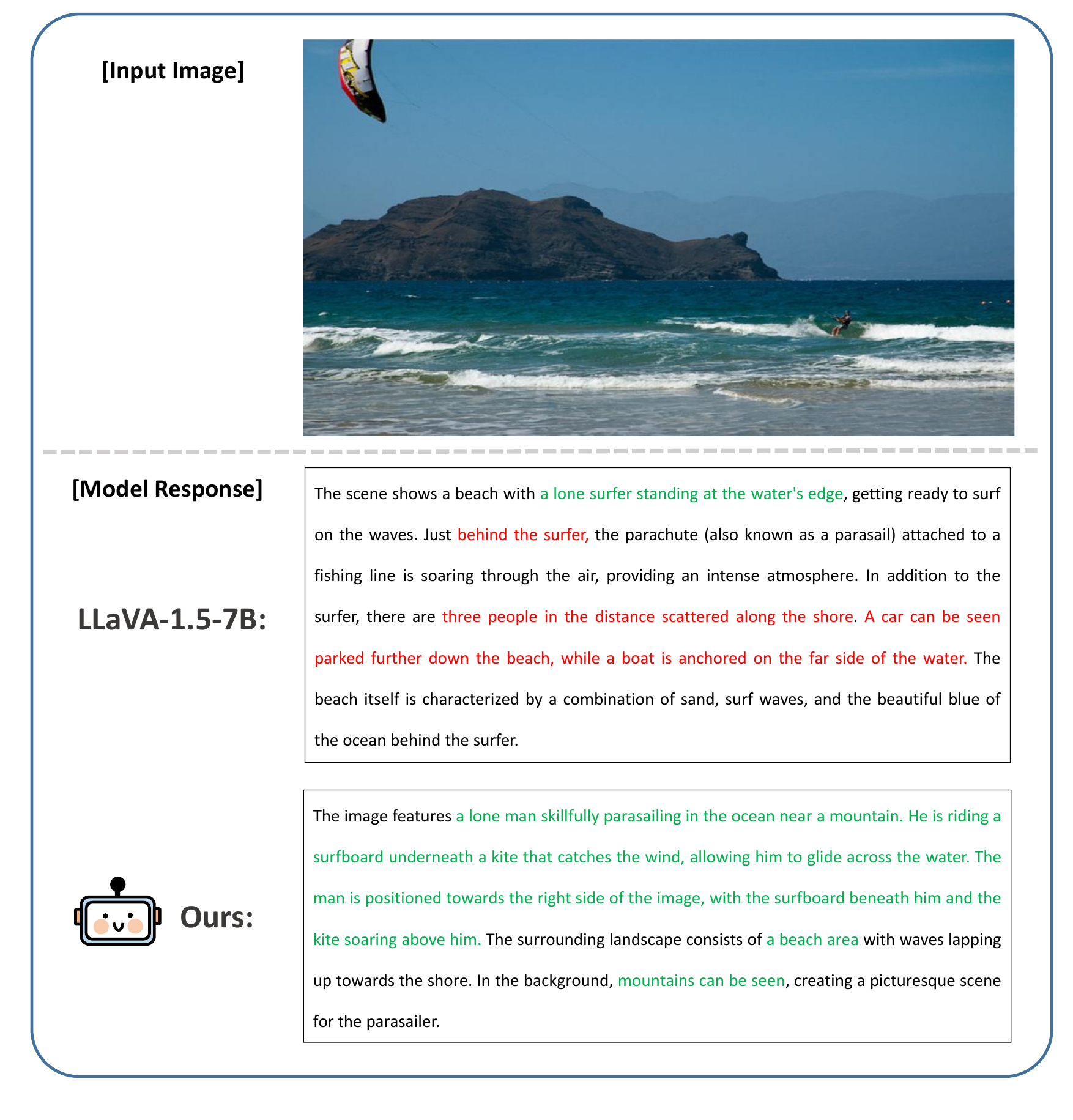}
\caption{One of the cases illustrates the effectiveness of our method. We use the prompt ``Please describe the image in detail.'' to get the caption.}
\label{fig:case_study2}
\end{figure*}

\begin{figure*}[t]
\centering
\includegraphics[width=0.8\linewidth]{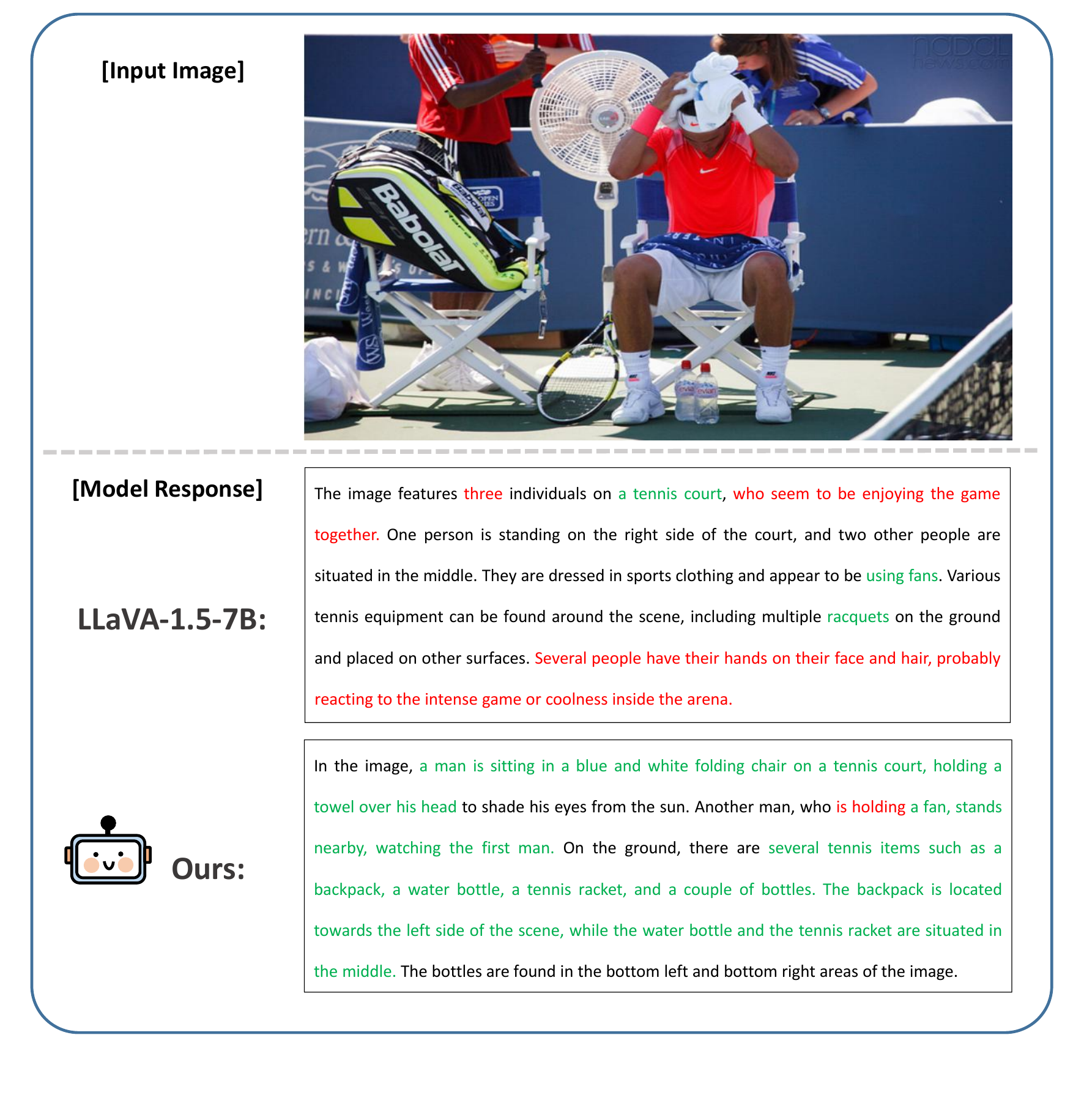}
\caption{One of the cases illustrates the effectiveness of our method. We use the prompt ``Please describe the image in detail.'' to get the caption.}
\label{fig:case_study3}
\end{figure*}

To show how our approach performs in open-ended captioning tasks, we offer several specific cases, as shown in Fig.\ref{fig:case_study1}, Fig.\ref{fig:case_study2}, and Fig.\ref{fig:case_study3}. As expected, the results show that our approach enables the LVLM to focus more on image regions and capture more specific visual elements. In contrast, the original model tends to describe the image from the overall layout and atmosphere, introducing hallucinated objects and lacking details. This finding keeps consistent with our quantitative CHAIR evaluation result.

\end{document}